\documentclass[runningheads]{llncs}
\usepackage{graphicx}
\begin{document}
\title{Large-Scale Pedestrian Retrieval Competition}
%\author{First Author\inst{1}\orcidID{0000-1111-2222-3333} \and
%Second Author\inst{2,3}\orcidID{1111-2222-3333-4444} \and
%Third Author\inst{3}\orcidID{2222--3333-4444-5555}}

\author{Da Li \and Zhang Zhang}

\authorrunning{D. Li et al.}
%\institute{Princeton University, Princeton NJ 08544, USA \and
%Springer Heidelberg, Tiergartenstr. 17, 69121 Heidelberg, Germany

\institute{}

\maketitle              % typeset the header of the contribution
\newcommand{\tabincell}[2]{\begin{tabular}{@{}#1@{}}#2\end{tabular}}
\section{Overview and Motivation}

The Large-Scale Pedestrian Retrieval Competition (LSPRC) mainly focuses on person retrieval which is an important end application in intelligent vision system of surveillance. Person retrieval aims at searching the interested target with specific visual attributes or images. The low image quality, various camera viewpoints, large pose variations and occlusions in real scenes make it a challenge problem.

By providing large-scale surveillance data in real scene and standard evaluation methods that are closer to real application, the competition aims to improve the robust of related algorithms and further meet the complicated situations in real application. LSPRC includes two kinds of tasks, i.e., Attribute based Pedestrian Retrieval (PR-A) and Re-IDentification (ReID) based Pedestrian Retrieval (PR-ID). The normal evaluation index, i.e., mean Average Precision (mAP), is used to measure the performances of the two tasks under various scale, pose and occlusion. While the method of system evaluation is introduced to evaluate the person retrieval system in which the related algorithms of the two tasks are integrated into a large-scale video parsing platform (named ISEE) combing with algorithm of pedestrian detection.

The competition is hold in Chinese Conference on Pattern Recognition and Computer Vision in 2018 (PRCV2018)\footnote{http://prcv.qyhw.net.cn/} and attract more than thirty institutions to participate. This report gives a brief analysis and conclusion on the competition results which may help researchers to develop more robust algorithms.

\section{Dataset}

\begin{figure}[!t]
  \centering
  \includegraphics[width=4.5in]{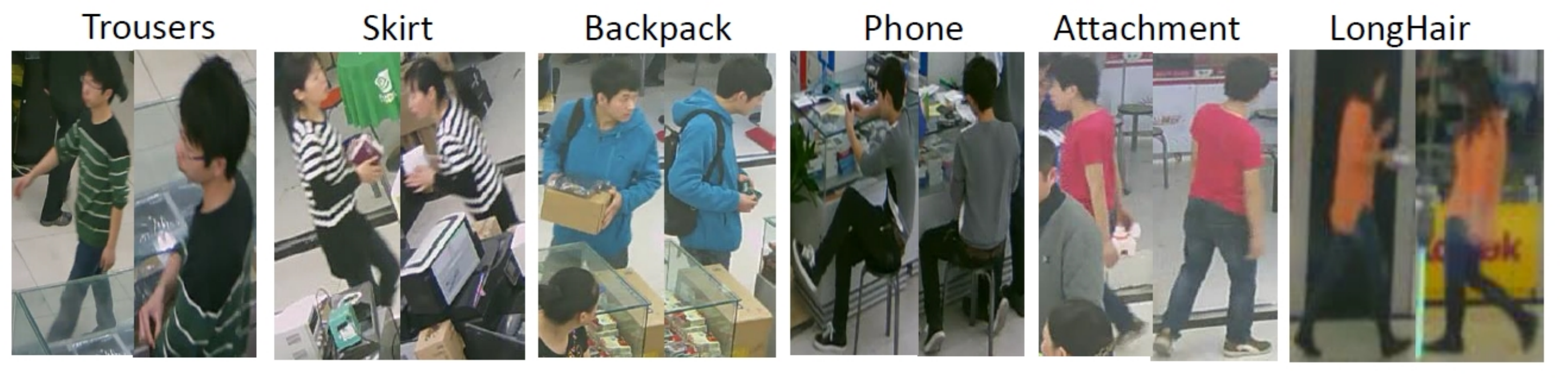}
  \caption{Image samples in RAP. In real scene, attributes change a lot due to camera viewpoint, body part occlusion, human pose, time range, and image quality, etc. These make the person retrieval a challenging problem.}
  \label{fig:RAP-examples}
\end{figure}

A richly annotated pedestrian (RAP) dataset \cite{iSEE:RAP2016}, which is collected for person retrieval in real visual surveillance scenarios, is adopted in LSPRC. A subset of samples is selected from RAP in this competition, which includes more than 68 thousands pedestrian images annotated with 72 fine-grade attributes and 2,589 identities (IDs). These samples are split into three parts, i.e., training, test and validation. Some basic information is listed in Table~\ref{table:RAP-LSPR}. And several samples in RAP are shown in Fig.~\ref{fig:RAP-examples}.

\begin{table*}[!hbt]
  \caption{Basic information of the data used in LSPRC.}
  \vspace{-0.2cm}
  \scriptsize
  \label{table:RAP-LSPR}
  \centering
  \begin{tabular}{|c|c|c|c|c|c|c|c|}
    \hline
    \multicolumn{4}{|c|}{Attribute} & \multicolumn{2}{c|}{Re-Identification} & \multicolumn{2}{c|}{Other Information} \\
    \hline
    \hline
     \tabincell{c}{\#Training \\Samples} & \tabincell{c}{\#Validation\\ Samples} & \tabincell{c}{\#Test \\Samples} & \#Attributes & \tabincell{c}{\#Training IDs \\ (samples)} & \tabincell{c}{\#Test IDs \\(samples)} & \tabincell{c}{Resolution \\($w \times h$)} & \# Cameras \\
    \hline
     33268 & 8317 & 25986 & 72 & \tabincell{c}{1295\\(13178)} & \tabincell{c}{1294\\{13460}} & \tabincell{c}{from 33$\times$81 \\to 415$\times$583} & 25 \\
    \hline
  \end{tabular}
\end{table*}

Besides the annotated RAP dataset, about 350 hours(h) raw HD videos corresponding to the labeled image samples are also provided for the system evaluations on the performances of Atributes-based or Re-Identification (ReID)-based pedestrian retrieval.

Moreover, another subset of whole image samples in RAP, which are annotated with pedestrian bounding boxes, are provided to train the person detection model that can be combined with subsequent recognition algorithms in the stage of system evaluation.

\section{Tasks}

Based on the different querying conditions, LSPRC includes two kinds of tasks, i.e., Attribute-based Pedestrian Retrieval (PR-A) and ReID-based Pedestrian Retrieval (PR-ID).

\subsection{Task of PR-A}
The task of PR-A in LSPRC can be seen as analogous to Attribute Recognition. The confidence scores of existed attributes are the basis for further retrieval. It worth to note that the model is employed to PR-A can only be trained with the training set in RAP.

PR-A can be divided into two stage based on the two kinds of evaluation methods, i.e., PR-A-RAP and PR-A-SYS.

\textbf{PR-A-RAP}. A list of query conditions are generated with different number of attributes (ranging from 1 to 4). Under different conditions, search the samples in test set based on the confidence scores which are obtained through attribute recognition.

\textbf{PR-A-SYS}. In this stage, different pipelines, through combing the submitted algorithms of attribute recognition with pedestrian detection, are generate to parse the 100h videos on ISEE. Meanwhile, inspired by the works \cite{iSEE:VTT4CVSystem} and \cite{iSEE:VTT2015}, a question-answering (QA) paradigm is designed to evaluate the performance of person retrieval system. As shown in Table~\ref{table:QUERY-LSPR}, more than 5 million polar (binary) queries (whether a person with the specified attribute(s)) are generated based on the ground truth of test set in RAP. The performance of the question-answering results reflet the performance of submitted algorithm integrated into the person retrieval system.

\subsection{Task of PR-ID}
The task of PR-A in LSPRC can be seen as analogous to person ReID. The trained ReID model is used to extract features for both the probe and gallery images. The features are further used to calculate the similarity between the probe and gallery image. PR-ID can be divided into two stage as well, i.e., PR-ID-RAP and PR-ID-SYS. No other auxiliary data related to person ReID is allowed in training.

\textbf{PR-ID-RAP}. A subset of samples in the test set is selected as the query set. Then, for each image in the query set, calculate the similarity with all the images in test set based on the ReID features. Sort all the images in test set descending depended on the similarity which is the basis to determine if the two samples the same ID or not.

\textbf{PR-ID-SYS}. Similar with PR-A-SYS, the submitted algorithms for ReID are combined with pedestrian detection to generate different pipelines to parse another 250h videos on ISEE.
Then, the KNN-search is conducted for all the detected pedestrians to get their top-K neighbours using ReID features. A relationship graph is further constructed with the results of KNN-search, where each node represents a person, and the edge reflects if the two nodes belong to the same ID. Meanwhile, more than 15 million polar queries (whether the two pedestrians with the same ID or not) are generated based on the ground truth of test set in RAP (see Table~\ref{table:QUERY-LSPR}). The performance of the question-answering results reflet the performance of submitted algorithm integrated into the person retrieval system.

\begin{table}[!hbt]
  \caption{Number of queries used in PR-A-SYS and PR-ID-SYS.}
  \vspace{-0.2cm}
  %\scriptsize
  \label{table:QUERY-LSPR}
  \centering
  \begin{tabular}{|c|c|c|c|}
    \hline
    \multicolumn{2}{|c|}{PR-A-SYS} & \multicolumn{2}{c|}{PR-ID-SYS} \\
    \hline
    \hline
     \tabincell{c}{\# Positive Queries} & \tabincell{c}{\# Negative Queries} & \tabincell{c}{\# Positive Queries} & \# Negative Queries \\
    \hline
     354,700 & 4,972,430 & 491,518 & 15,723,652 \\
    \hline
  \end{tabular}
\end{table}

\section{Evaluation}

More than fifteen valid models about the two tasks are submitted finally. The evaluation results of PR-A are list in Table~\ref{table:LSPR-res-attr}, in which DeepMAR \cite{iSEE:DeepMAR2015} is the baseline model which is fine-tuned with the training data provided by LSPRC. We can find that the performances of the champion model on both PR-A-RAP and PR-A-SYS are inferior to those of the baseline method. It indicates that the task of attribute recognition is still a challenge problem in real surveillance scenes.

\begin{table}[h]
  \caption{The evaluation results of PR-A.}
  \vspace{-0.2cm}
  \renewcommand\tabcolsep{1pt}
  \scriptsize
  \label{table:LSPR-res-attr}
  \centering
  \begin{tabular}{|c|l|c|c|}
    \hline
     Team & Method & \tabincell{c}{mAP\\(PR-A-RAP)} & \tabincell{c}{$F_1$ score\\(PR-A-SYS)} \\
    \hline
    \hline
     Deeplning & \tabincell{l}{Modify the conv-layer in the attention module \\of \cite{iSEE:DIAC2018} with deformable convolution \cite{iSEE:DeformableCNN2017}.} & \textbf{0.4220} & \textbf{0.4135} \\
    \hline
     ASTRI     & \tabincell{l}{Densenet (Backbone) for features extraction; \\another four branches for attributes prediction.} & 0.4107 & 0.2042 \\
    \hline
     Xiangtan University & \tabincell{l}{VPN\cite{iSEE:VesPA2017}: train a classifier for view point;\\ SRN\cite{iSEE:SRN2017}: learn the multi-label relations \\among images.} & 0.3512 & 0.2455 \\
    \hline
     TYUT      & Xception \cite{iSEE:xception2017} & 0.0777 & 0.0019 \\
    \hline
    \hline
     DeepMAR \cite{iSEE:DeepMAR2015} & \tabincell{l}{Multi-label classification with ResNet50.} & 0.4267 & 0.4196 \\
    \hline
  \end{tabular}
\end{table}

Table~\ref{table:LSPR-res-reid} shows the evaluation results on PR-ID. MSCAN \cite{iSEE:MSCAN2017} listed in the last row is the baseline method. The value of its performance (mAP) is copied from the original paper directly. And we also did not evaluate it in the stage of PR-ID-SYS. Different from the results in PR-A, most of the submitted models achieve superior performance than baseline method on PR-ID-RAP. Moreover, from the column of \emph{Method}, we can find that multi-model fusion and part-based model are two common strategy to improve the performances of PR-ID.

\begin{table}[h]
  \caption{The evaluation results of PR-ID.}
  \vspace{-0.2cm}
  \renewcommand\tabcolsep{1pt}
  \scriptsize
  \label{table:LSPR-res-reid}
  \centering
  \begin{tabular}{|c|l|c|c|}
    \hline
     Team & Method & \tabincell{c}{mAP\\(PR-ID-RAP)} & \tabincell{c}{$F_1$ score\\(PR-ID-SYS)} \\
    \hline
    \hline
     Dahua Technology & \tabincell{l}{Multi-model fusion: \\~~~~~~~PCB \cite{iSEE:PCB2018} \& MGN \cite{iSEE:MGN2018}\\ Backbone: \\~~~~~~~seresnet152, resnet152, densenet201} & \textbf{0.7335} & \textbf{0.5286} \\
    \hline
     SYSU \& ZNV Technology &  \tabincell{l}{Multi-model fusion: PCB \cite{iSEE:PCB2018} \& MGN \cite{iSEE:MGN2018}\\ Backbone: resnet101, senet50, densenet129} & 0.7087          & 0.5263 \\
    \hline
     Weihua Chen &   \tabincell{l}{Multi-model fusion \& \\human parsing assistant;\\ Local feature learning: \\~~~~~~~PCB \cite{iSEE:PCB2018} \& MGN \cite{iSEE:MGN2018}; \\ Global feature learning: midfeat \cite{iSEE:devil2017}}  & 0.6421          & 0.5030 \\
    \hline
     \tabincell{c}{The Army Engineering\\ University of PLA}  & Multi-model fusion  & 0.6059          & 0.4802 \\
    \hline
     DiDi Research &  MGN \cite{iSEE:MGN2018} & 0.5933 & 0.4906 \\
    \hline
     \tabincell{c}{ZJU \& ZJUT \\ \& iCareVision} &   -           & 0.5182 & 0.4629 \\
    \hline
     Wave Kingdom &         -                  & 0.4984 & 0.3862 \\
    \hline
     deeplning &            -                   & 0.4601 & 0.3844 \\
    \hline
     \tabincell{c}{China University \\of Petroleum} &     -      & 0.4542 & 0.3852\\
    \hline
     SYSU-NSCC &               -                & 0.4492 & 0.4108 \\
    \hline
     ASTRI &                  -                 & 0.3951 & 0.3242\\
    \hline
     TYUT &                  -                 & 0.1461 & 0.1618 \\
    \hline
     Anhui University &       -                 & - & 0.2393 \\
    \hline
     GDUT &                  -                  & - & 0.1375\\
    \hline
    \hline
     MSCAN \cite{iSEE:MSCAN2017}& \tabincell{l}{Cut the pedestrian image into \\several parts softly.} & 0.3828 & - \\
    \hline
  \end{tabular}
\end{table}

\section{Conclusion}

In this report, we introduced the Large-Scale Pedestrian Retrieval Competition (LSPRC) including the dataset, tasks and evaluation results. From the results, we can summarize as follows.
\begin{itemize}
  \item The best performances of the two tasks are lower than 60\%, which are not satisfying for real applications. So person retrieval is still a challenge problem.
  \item For the task of PR-ID, multi-model fusion and part-based model are the most common strategies in submitted methods. Especially, the local features can be extracted through the part-based model, which provide more fine-grained information about spatial alignment. So the methods, such as DensePose \cite{Guler2018DensePose}, which can provide more precision information on alignment, have attracted extensive attention on developing modern ReID models.
\end{itemize}

\bibliographystyle{splncs04}
\bibliography{iSEE}

\begin{thebibliography}{10}
\providecommand{\url}[1]{\texttt{#1}}
\providecommand{\urlprefix}{URL }
\providecommand{\doi}[1]{https://doi.org/#1}

\bibitem{iSEE:xception2017}
Chollet, F.: Xception: Deep learning with depthwise separable convolutions. In:
  computer vision and pattern recognition. pp. 1800--1807 (2017)

\bibitem{iSEE:DeformableCNN2017}
Dai, J., Qi, H., Xiong, Y., Li, Y., Zhang, G., Hu, H., Wei, Y.: Deformable
  convolutional networks. In: international conference on computer vision. pp.
  764--773 (2017)

\bibitem{iSEE:VTT4CVSystem}
Geman, D., Geman, S., Hallonquist, N., Younes, L.: Visual turing test for
  computer vision systems. Proceedings of the National Academy of Sciences
  \textbf{112}(12),  3618--3623 (2015)

\bibitem{Guler2018DensePose}
Guler, R.A., Neverova, N., Kokkinos, I.: Densepose: Dense human pose estimation
  in the wild. In: CVPR (2018)

\bibitem{iSEE:DeepMAR2015}
Li, D., Chen, X., Huang, K.: Multi-attribute learning for pedestrian attribute
  recognition in surveillance scenarios. In: 2015 3rd IAPR Asian Conference on
  Pattern Recognition. pp. 111--115 (2015)

\bibitem{iSEE:MSCAN2017}
Li, D., Chen, X., Zhang, Z., Huang, K.: Learning deep context-aware features
  over body and latent parts for pedestrain re-identification. In: Proc. {IEEE}
  International Conference on Computer Vision and Pattern Recognition. pp.
  384--393 (2017)

\bibitem{iSEE:RAP2016}
Li, D., Zhang, Z., Chen, X., Huang, K.: A richly annotated pedestrian dataset
  for person retrieval in real surveillance scenarios. IEEE Transactions on
  Image Processing  \textbf{28}(4),  1575--1590 (2019)

\bibitem{iSEE:VTT2015}
Qi, H., Wu, T., Lee, M.W., Zhu, S.C.: A restricted visual turing test for deep
  scene and event understanding. arXiv preprint arXiv:1512.01715  (2015)

\bibitem{iSEE:DIAC2018}
Sarafianos, N., Xu, X., Kakadiaris, I.A.: Deep imbalanced attribute
  classification using visual attention aggregation. In: european conference on
  computer vision. pp. 708--725 (2018)

\bibitem{iSEE:VesPA2017}
Sarfraz, M.S., Schumann, A., Wang, Y., Stiefelhagen, R.: Deep view-sensitive
  pedestrian attribute inference in an end-to-end model. In: british machine
  vision conference (2017)

\bibitem{iSEE:PCB2018}
Sun, Y., Zheng, L., Yang, Y., Tian, Q., Wang, S.: Beyond part models: Person
  retrieval with refined part pooling (and a strong convolutional baseline. In:
  european conference on computer vision. pp. 501--518 (2018)

\bibitem{iSEE:MGN2018}
Wang, G., Yuan, Y., Chen, X., Li, J., Zhou, X.: Learning discriminative
  features with multiple granularities for person re-identification. In: acm
  multimedia. pp. 274--282 (2018)

\bibitem{iSEE:devil2017}
Yu, Q., Chang, X., Song, Y., Xiang, T., Hospedales, T.M.: The devil is in the
  middle: Exploiting mid-level representations for cross-domain instance
  matching. arXiv: Computer Vision and Pattern Recognition  (2017)

\bibitem{iSEE:SRN2017}
Zhu, F., Li, H., Ouyang, W., Yu, N., Wang, X.: Learning spatial regularization
  with image-level supervisions for multi-label image classification. In:
  computer vision and pattern recognition. pp. 2027--2036 (2017)

\end{thebibliography}
%Author, F.: Article title. Journal \textbf{2}(5), 99--110 (2016)
%Author, F., Author, S.: Title of a proceedings paper. In: Editor,
%F., Editor, S. (eds.) CONFERENCE 2016, LNCS, vol. 9999, pp. 1--13.
%Springer, Heidelberg (2016). \doi{10.10007/1234567890}
%Author, F., Author, S., Author, T.: Book title. 2nd edn. Publisher,
%Author, A.-B.: Contribution title. In: 9th International Proceedings
%on Proceedings, pp. 1--2. Publisher, Location (2010)
%LNCS Homepage, \url{http://www.springer.com/lncs}. Last accessed 4
%\end{thebibliography}
\end{document}